\documentclass[10pt,twocolumn,letterpaper]{article}

\usepackage[pagenumbers]{iccv} 

\definecolor{iccvblue}{rgb}{0.21,0.49,0.74}
\usepackage[pagebackref,breaklinks,colorlinks,allcolors=iccvblue]{hyperref}

\def\paperID{4605} 
\def\confName{ICCV}
\def\confYear{2025}

\usepackage{capt-of}
\usepackage[dvipsnames]{xcolor}
\usepackage{xspace}
\usepackage{enumitem}
\usepackage{amsmath,amssymb,amsbsy,amsfonts,dsfont,pifont,bm,bbm,mathrsfs,mathtools,nicefrac}
\usepackage{algorithm,algpseudocode,listings}
\usepackage{booktabs,multirow,adjustbox,diagbox,threeparttable,tabularray,colortbl}
\usepackage{bbding}
\usepackage{indentfirst} 

\newcommand{\method}{\texttt{DanceLCM}\xspace}

\definecolor{Gray}{gray}{0.94}
\definecolor{liGray}{gray}{0.5}
\definecolor{LightCyan}{rgb}{0.88,1,1}

\usepackage{colortbl} 
\usepackage{xcolor}

\definecolor{cvprblue}{rgb}{0.21,0.49,0.74}
\usepackage{wrapfig}
\usepackage[capitalize]{cleveref}  
\crefname{section}{Sec.}{Secs.}
\Crefname{section}{Section}{Sections}
\crefname{table}{Tab.}{Tabs.}
\Crefname{table}{Table}{Tables}
\crefname{figure}{Fig.}{Figs.}
\Crefname{figure}{Figure}{Figures}
\crefname{equation}{Eq.}{Eqs.}
\Crefname{equation}{Equation}{Equations}
\newlength\savewidth\newcommand\shline{\noalign{\global\savewidth\arrayrulewidth
  \global\arrayrulewidth 1pt}\hline\noalign{\global\arrayrulewidth\savewidth}}

\title{Taming Consistency Distillation for Accelerated Human Image Animation}

\author{
   \hspace{-0.4cm} 
   Xiang Wang$^{1}$
     \hspace{0.01cm} 
    Shiwei Zhang$^{2}$
     \hspace{0.01cm} 
    Hangjie Yuan$^{2}$
     \hspace{0.01cm} 
    Yujie Wei$^{3}$
     \hspace{0.01cm} 
    Yingya Zhang$^2$ 
     \hspace{0.01cm}
    Changxin Gao$^1$ \\
    \hspace{-0.4cm} 
    Yuehuan Wang$^1$ 
     \hspace{0.01cm}
     Nong Sang$^{1}$
      \vspace{2mm}
     \\
    $^1$Key Laboratory of Image Processing and Intelligent Control,\\   School of Artificial Intelligence and Automation, Huazhong University of Science and Technology\\
      $^2$Alibaba Group \hspace{0.5cm} $^3$Fudan University 
}

\begin{document}

\twocolumn[{
\maketitle
    \centering
    \vspace{-15pt}
    \includegraphics[width=1.0\textwidth]{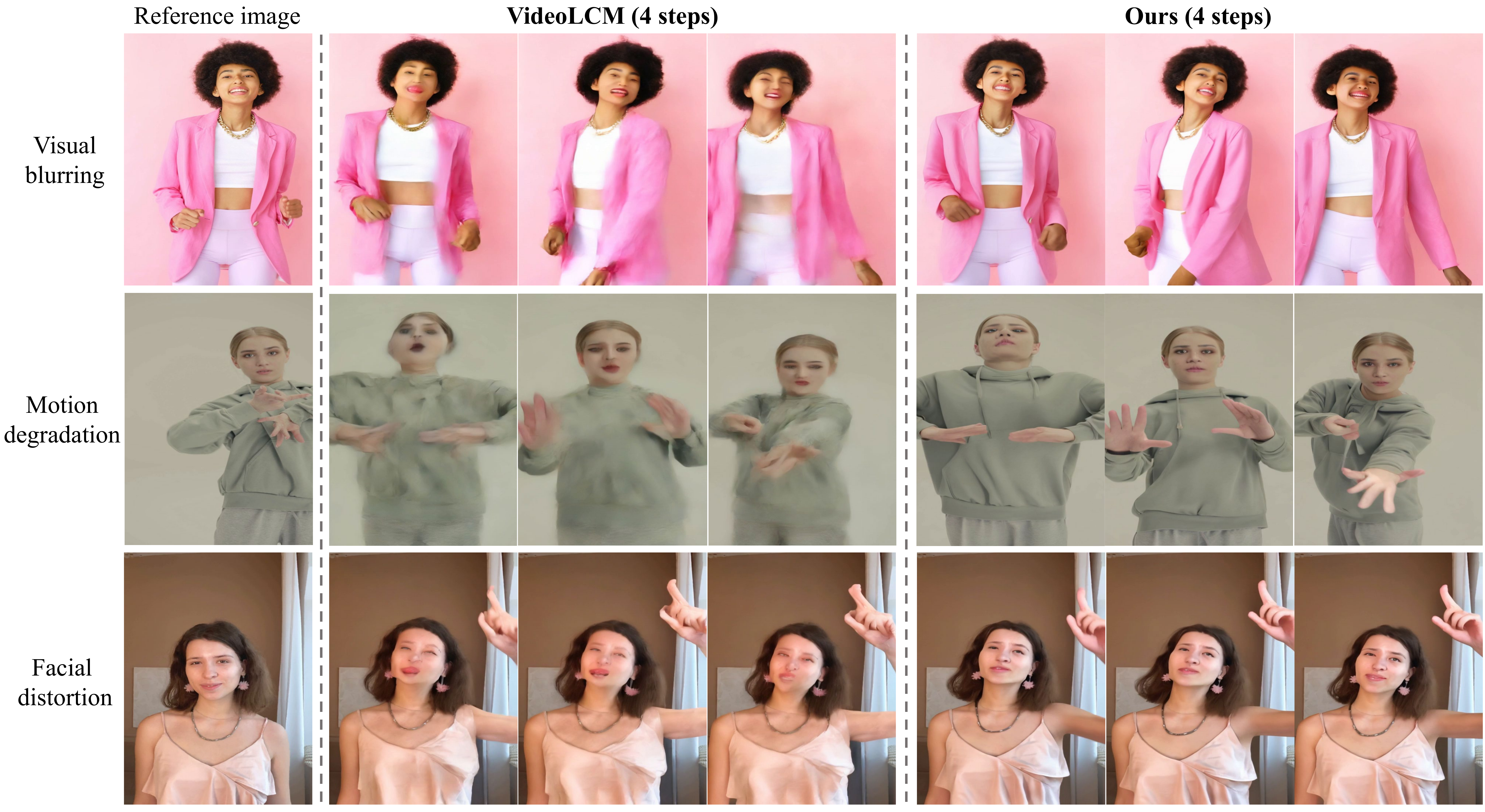}
    \vspace{-21pt}
    \captionof{figure}{
        {Comparative results} generated by the baseline VideoLCM~\cite{wang2023videolcm} and our proposed \method on human image animation task.
    }
    \label{teaser}
    \vspace{10pt}
    }
]

\begin{abstract}

Recent advancements in human image animation have been propelled by video diffusion models, yet their reliance on numerous iterative denoising steps results in high inference costs and slow speeds.
An intuitive solution involves adopting consistency models, which serve as an effective acceleration paradigm through consistency distillation. However, simply employing this strategy in human image animation often leads to quality decline, including visual blurring, motion degradation, and facial distortion, particularly in dynamic regions.
In this paper, we propose the \method approach complemented by several enhancements to improve visual quality and motion continuity at low-step regime:
(1) segmented consistency distillation with an auxiliary light-weight head to incorporate supervision from real video latents,
mitigating cumulative errors resulting from single full-trajectory generation;
(2) a motion-focused loss to centre on motion regions, and explicit injection of facial fidelity features to improve face authenticity.
Extensive qualitative and quantitative experiments demonstrate that \method achieves results comparable to state-of-the-art video diffusion models with a mere 2-4 inference steps, significantly reducing the inference burden without compromising video quality.
{
The code and models will be made publicly available.
}

\end{abstract}

\vspace{-10pt}
\section{Introduction}
\label{sec:intro}

%

Human image animation~\cite{Animateanyone,magicanimate,champ,wang2024unianimate,zhang2024mimicmotion,peng2024controlnext} aims to generate temporally coherent video content conditioned on a reference image and a human pose sequence and has witnessed remarkable advancements with the advent of video diffusion models~\cite{modelscopet2v,guo2023animatediff,videocomposer,VideoLDM,tft2v}.
Recent approaches, such as Animate Anyone~\cite{Animateanyone} and UniAnimate~\cite{wang2024unianimate}, have successfully extended 2D diffusion models~\cite{stablediffusion,DDIM,DDPM,EvolveDirector} into the video domain by integrating additional temporal layers, thus effectively modeling spatiotemporal dependencies and achieving impressive results. However, the inherent requirement of diffusion models for numerous denoising steps has led to high computational costs and slow inference speeds, serving as a bottleneck for practical applications.
%

To bridge this gap,
consistency models~\cite{song2023consistency,luo2023latent} have emerged as a promising acceleration technique, demonstrating remarkable success in various generation fields~\cite{VideoLDM,luo2023latent,kim2023consistency,heek2024multistep,wang2024animatelcm}. By employing consistency distillation, these models constrain any adjacent points along the probabilistic flow ODE (PF-ODE) trajectory to map to the same origin. In the realm of video generation, pioneering works such as VideoLCM~\cite{wang2023videolcm} 
leverage consistency distillation strategy to approach the performance of teacher video diffusion models and support fast video synthesis with about 4 steps.
Despite these efforts, directly extending consistency distillation to the task of human image animation leads to visual blurring in dynamic regions, motion degradation, and facial distortion, as depicted in~\cref{teaser}. 
This is primarily due to the inherent difficulty in constraining any point on a PF-ODE trajectory to be mapped back to the trajectory's starting point~\cite{heek2024multistep,xie2024mlcm,wang2024phased}, overlooking the unique characteristics inherent to different phase of the trajectory~\cite{qian2024boosting,hu2024ella}. For instance, the early denoising stage prioritizes low-frequency semantics, while the later stage emphasizes high-frequency details, leading to significant learning difficulty.
%
To address this, we introduce the segmented trajectory distillation technique~\cite{heek2024multistep,xie2024mlcm} to this task, which redefines the consistency mapping as a segmented mapping that partially alleviates these issues, 
but it still fails to achieve an optimal model.
%
The key challenges lie in the following two critical aspects:
\textbf{1)} \textit{Suboptimal distillation supervision}: 
During consistency distillation, the output of the student model is forced to be aligned with its exponentially moving average (EMA) counterpart.
%
However, distillation will limit the quality of the generated videos to that of the teacher diffusion model.
The supervision signals derived from the teacher model are suboptimal due to the typically poor generative quality exhibited by video diffusion models, which consequently propagate errors to the student model.
\textbf{2)} \textit{Minimal focus on motion}: 
The distilled model tends to generate videos with motion blur and distortion due to the increased difficulty of optimizing higher-order complex human motion dynamics in a limited number of steps, particularly in facial regions that are highly sensitive to human perception.

%

%

%
In this work, we propose a novel approach \method built upon on consistency distillation, enhanced with several techniques to improve distillation supervision and increase focus on motion.
Specifically, we introduce segmented trajectory distillation for human image animation, which reduces optimization difficulty and error accumulation by segmenting the PF-ODE trajectory rather than generating it all at once. 
To provide more reliable distillation supervision, we incorporate an auxiliary head that aligns predicted videos with real video latents.
On the other hand,
to alleviate motion degradation and blurring in dynamic areas, we identify dynamic regions with minimal effort, enabling targeted focus on movable regions.
Furthermore, to preserve the fidelity of human faces, we integrate face-centered features to enhance facial realism. 
We conducted extensive qualitative and quantitative experiments on the standard TikTok~\cite{tiktokdata} and UBC Fashion~\cite{UBCfashion} datasets, demonstrating the effectiveness of the proposed \method.
\section{Related Work}
\label{sec:related_work}

This work intersects prominently with two related areas: human image animation and consistency distillation. 

\vspace{1mm}
\noindent \textbf{Human image animation.}
Given a reference image and a series of human poses, this task focuses on synthesizing photorealistic videos adhering to the input conditions~\cite{magicdance,champ,zhu2024poseanimate,chang2023muse,zhang2024mimicmotion,peng2024controlnext,FOMM,li2019dense,karras2023dreampose,xu2024you,tan2024animatex,zhou2024dormant,ma2024follow,shao2024human4dit,wang2024humanvid,zhou2024realisdance,wang2024vividpose,TPS,MRAA}.
%
Disco~\cite{disco} introduces a decomposition approach for input conditions, effectively allowing for separate control of background and foreground generation. 
AnimateAnyone~\cite{Animateanyone} and MagicAnimate~\cite{magicanimate} employ video diffusion models in human image animation, utilizing an auxiliary reference network to encapsulate the appearance attributes of the reference image. 
Champ~\cite{champ} investigates the complementary relationships between multiple pose features, further enriching the generation process.
UniAnimate~\cite{wang2024unianimate} presents a unified video diffusion model, combining appearance alignment with motion control through supplementary reference pose information, thus facilitating enhanced appearance alignment.
%
Despite these advancements, existing methods typically require extensive inference denoising steps to achieve satisfactory outputs, leading to considerable resource consumption and slow speeds.
To this end, 
this work attempts to investigate the accelerated human image animation technique without losing video quality.
%

\begin{figure*}[t]
    \centering
    \includegraphics[width=0.99\linewidth]{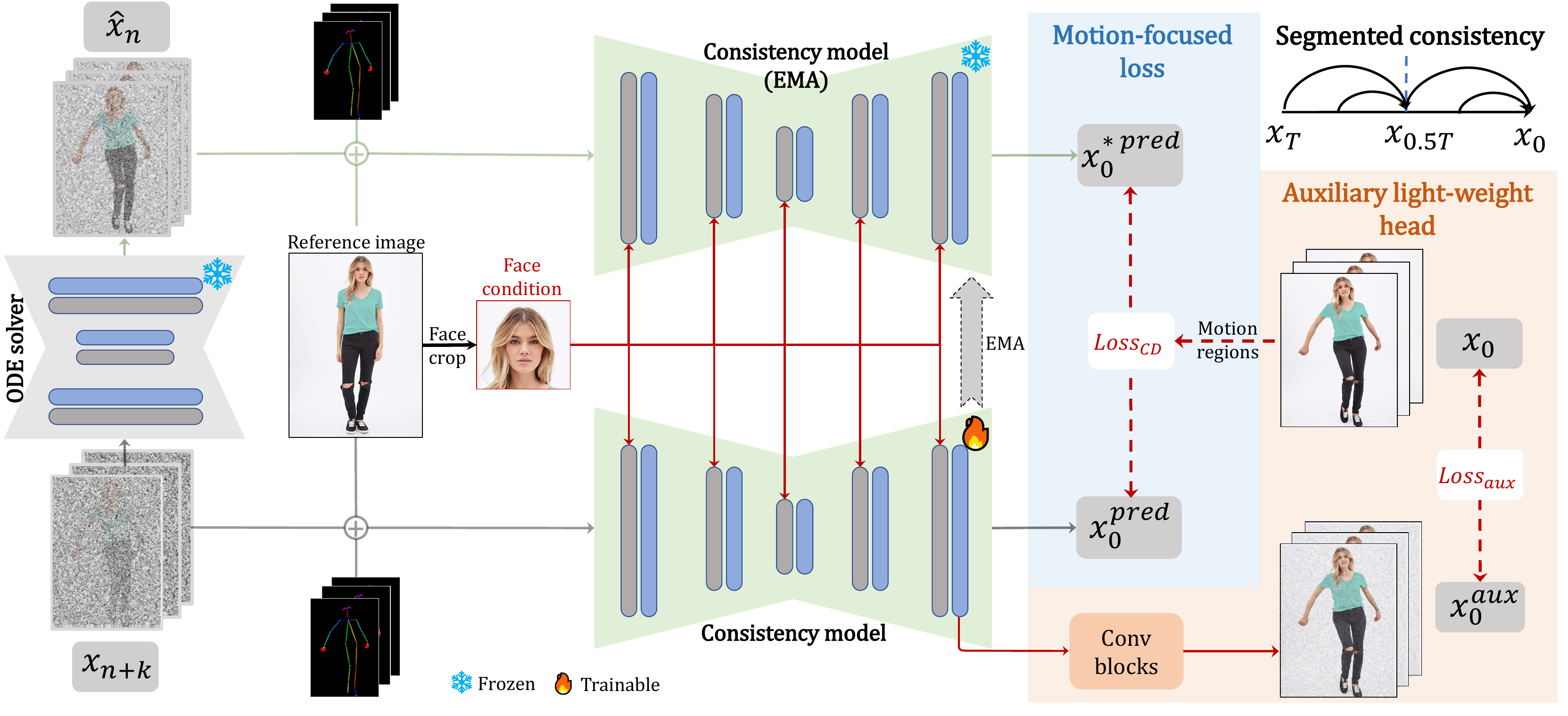}
    \vspace{-5mm}
    \caption{Overall pipeline of the proposed \method. 
    The segmented trajectory distillation is performed to transfer the knowledge of the pretrained teacher diffusion model (\ie, the ODE solver) to the student consistency model by forcing the outputs of two consistency models to be consistent.
    Furthermore, an auxiliary loss that aligns predicted video latents with real video latents is adopted  to provide more reliable distillation supervision.
    Additionally, a motion-focused loss is applied to emphasize motion regions, and the facial condition is explicitly injected into the model to improve facial realism.
    }
    \label{fig:Network}
    \vspace{-2mm}
\end{figure*}

\vspace{1mm}
\noindent \textbf{Consistency distillation.}
The intrinsic requirement for extensive iterative denoising steps in diffusion models has fueled exploration into various acceleration techniques~\cite{xie2024mlcm,lin2024sdxl,luo2023latent,kong2023act,ren2024hyper,xu2024ufogen,liu2023instaflow,yin2024one,yan2024perflow,sauer2024fast,zheng2024trajectory,zhang2024sf,mao2024osv}.
Among these, the consistency model~\cite{song2023consistency,luo2023latent,song2023improved} has emerged as a prominent algorithm due to its stable training regime and promising results, enforcing self-consistency property along the PF-ODE trajectory.
%
%
Subsequent works, such as LCM~\cite{luo2023latent}, attempt to extend consistency distillation from pixel space to latent space while introducing classifier-free guidance coefficients to enhance fidelity.
Multistep consistency model~\cite{heek2024multistep} innovatively segments the ODE trajectory into  several equidistant sub-trajectories, imposing self-consistency characteristics within each sub-trajectory, and facilitates a favorable trade-off between sampling speed and generation quality.
Some recent works~\cite{wang2024phased,kong2023act,zhang2024sf,mao2024osv} attempt to combine distillation techniques and generative adversarial networks (GANs)~\cite{goodfellow2020generative,skorokhodov2022stylegan} to improve details while suffering from the prohibitive computational
 cost and difficulty of convergence due to the training instability of GANs~\cite{salimans2016improved}.
In addition, consistency distillation retains the capability to iteratively refine results, which is a merit compared to existing one-step GAN-based approaches~\cite{mocogan,zhang2024sf,mao2024osv}.
Following these successes in the image domain, VideoLCM~\cite{wang2023videolcm} and AnimateLCM~\cite{wang2024animatelcm} started to extend consistency distillation into video generation realms.
In this work, we mainly focus on 
human image animation task and present several techniques for consistency distillation to improve the visual quality and temporal coherence.

\section{Method}

In this section, we first provide a brief introduction to the preliminaries of the video diffusion model and consistency model.
Then, we will elaborate on the detailed architecture of \method.
The overall pipeline of the proposed \method is illustrated in \cref{fig:Network}.

\subsection{Preliminaries}
\noindent
\textbf{Video diffusion model.}
Through incorporating a temporal dimension,
video diffusion models~\cite{VideoLDM,modelscopet2v,videocomposer,cogvideo,imagenvideo,make-a-video,wei2023dreamvideo,qiu2024freescale,yuan2023instructvideo,tft2v,zhang2023i2vgen,tune-a-video} extend the conventional 2D diffusion framework to generate coherent video sequences. Existing diffusion-based human image animation methods~\cite{magicanimate,wang2024unianimate} typically involve a series of denoising steps where noise is progressively reduced, allowing for the generation of high-fidelity video outputs that conform to a given reference image and pose motion.
The mathematical formulation of the video diffusion model can be expressed as follows:
\begin{equation}
    x_{t}=\sqrt{\alpha _{t}}x_{0}+ \sqrt{1-\alpha _{t}}\epsilon 
    \label{eq:1}
\end{equation}
where $x_{0}$ denotes the original clean video, $x_{t}$ is the noisy data at timestep $t$, $\alpha _{t}$ is a scheduling factor, and $\epsilon$ represents the Gaussian noise. The reverse process aims to step back from $x_{T}$ to $x_{0}$ via a learned denoising network~\cite{modelscopet2v,tft2v,VideoLDM,guo2023animatediff,wang2024unianimate,magicanimate},
and $T$ is a sufficiently large value and is usually set to 1000. The reverse process can be formulated as:
\begin{equation}
p_{{\theta}^{\prime} }(x_{t-1}|x_{t}) = \mathcal{N}(x_{t-1};\mu_{{\theta}^{\prime}}(x_{t},t),{\textstyle \sum_{{\theta}^{\prime}}}(x_{t},t) )
  \label{eq:2}
\end{equation}
where we usually train a 3D-UNet model parameterized by ${\theta}^{\prime}$  to approximate the clean data progressively.

\vspace{1mm}
\noindent
\textbf{Consistency distillation.} The
consistency model~\cite{song2023consistency,luo2023latent} facilitates accelerated denoising inference in generative processes through consistency distillation. 
The distillation process attempts to distill rich knowledge from a well-trained teacher diffusion model, enabling accelerated inference.
To achieve this goal, it enforces the self-consistency property between the outputs of nearby points along the PF-ODE trajectory.
Mathematically, the optimization objective is to minimize the following form:
\begin{equation}
\mathcal{L}_{CD} = d(x^{pred}_{0}, x^{* pred}_{0}) 
 =  d(f_{\theta }(x_{{n+k}}, {n+k}), f_{\theta^- }({\hat{x}}_{n}, n))
  \label{eq:3}
\end{equation}
where $\theta^-$ is derived from the exponential moving average (EMA) of parameters $\theta$ of the student consistency model, and ${\hat{x}}_{n}$  is the output estimated by the ODE solver, \ie, the teacher diffusion model, with $x_{{n+k}}$ as the noised input.
Refer to~\cite{luo2023latent} for more details about consistency distillation.

\subsection{DanceLCM}

Despite remarkable advancements of consistency models, naively applying consistency distillation to human image animation may result in issues such as human motion blur, degradation, and face distortion, primarily stemming from the difficulties associated with the full-trajectory distillation and the loss optimization mechanisms inherent in prior methods. To tackle these challenges, we propose a tailored consistency distillation methodology, \method, which integrates several key improvements.

\vspace{1mm}
\noindent \textbf{Segmented trajectory distillation with GT supervision.}
The conventional consistency distillation methods to learning a mapping function across the entire PF-ODE trajectory proves to be excessively complex~\cite{heek2024multistep,xie2024mlcm} and neglects the unique characteristics at different stages of the PF-ODE trajectory~\cite{qian2024boosting,hu2024ella}. The early denoising stage primarily focuses on low-frequency semantics, while the later stage emphasizes high-frequency details.
The characteristics at different stages lead to significant learning difficulty. To mitigate this, we introduce the segmented trajectory distillation strategy~\cite{heek2024multistep,xie2024mlcm}, where the PF-ODE trajectory is equally divided into $K$ segments.
%
We denote
the edge timesteps of these segments  as $s_{0}, s_{1}, ..., s_{K}$.
Self-consistency is enforced within each segment:
\begin{equation}
    \mathcal{L}_{CD} 
 =  d(f_{\theta }(x_{{t_m}}, {t_m}, s_{o}), f_{\theta^- }({\hat{x}}_{t_n}, t_n, s_{o}))
\end{equation}
where $t_{m} \in \mathcal{U}[s_{o},s_{o+1}]$, $t_{n} \in \mathcal{U}[s_{o},t_{m})$.
%
To further boost the visual fidelity of generated outputs, we introduce an additional light-weight head that directly aligns predicted video latents and their corresponding ground-truth latents. 
Using extra ground-truth video latents as supervision provides a more accurate reference, ensuring that the model captures finer temporal and spatial details, ultimately improving the quality and realism of the generated videos.
This loss is formulated as:
\begin{equation}
    \mathcal{L}oss_{aux} 
 =  d(x^{aux}_{0}, x_{0})
\end{equation}
where  $x^{aux}_{0}$ and  $x_{0}$ are the predicted and real video latents.

\vspace{1mm}
\noindent \textbf{Motion-focused loss.}
We empirically found that
the distilled model still results in output degradation to some extent, with human motion blur and facial distortion being the most noticeable artifacts to human perception.
To address the issue of human motion blur, we implement a motion-focused loss that prioritizes regions of significant motion. 
We calculate frame differences between consecutive frames to identify areas with prominent motion~\cite{wang2021tdn,wang2023molo}, 
while requiring negligible computational effort, mathematically defined as:
\begin{equation}
    \mathcal{M} 
 = \{(i,j)|[|v^{i,j}_{t}-v^{i,j}_{t+1}| > \delta ] \cup [|v^{i,j}_{t}-v^{i,j}_{t-1}| > \delta ] \} 
\end{equation}
where 
$\delta$ is a predefined threshold (\eg, 0.2) that indicates the degree of motion. 
In motion regions,
we assign increased loss weights during training:
{
\begin{footnotesize}
\begin{equation}
    \mathcal{L}oss_{CD} 
 = ||x_{0}^{pred}-x_{0}^{*pred}|| + \sum_{{(i,j)\in \mathcal{M}}}{\lambda}_{1} \cdot||x_{0}^{pred,i,j}-x_{0}^{*pred,i,j}||
\label{eq:7}
\end{equation}
\end{footnotesize}%
}%
where ${\lambda}_{1}$ is a factor, such as $0.5$.
This targeted approach allows the model to focus more on the challenging human motion areas, reducing  motion blur and artifacts.

\vspace{1mm}
\noindent \textbf{Facial fidelity enhancement.}
Due to the significant role that facial detail plays in human image animation, we explicitly incorporate facial fidelity features into \method. 
Specifically, we first detect the face part of the reference image and then use the variational autoencoder (VAE) encoder for feature extraction.
Finally, the VAE face feature is injected into the model by cross-attention after concatenating it together with the CLIP image feature of the reference image. 
The advantage of employing a VAE encoder to extract facial representation is its ability to prevent information loss. 
This ensures that the generated faces adhere to realistic attributes by consuming a rich set of facial feature representations extracted from the reference images.
In contrast, the CLIP encoder primarily captures general global semantic information, often missing finer local facial details~\cite{wang2024instantid,xing2024inv}.
%
%
%

In the training phase, we train our network end-to-end and optimize it using the overall loss described below:
\begin{equation}
    \mathcal{L}oss 
 = \mathcal{L}oss_{CD} + {\lambda}_{2} \cdot \mathcal{L}oss_{aux}
\label{eq:8}
\end{equation}
%
where ${\lambda}_{2}$ is a balancing coefficient, empirically set to $0.1$.
During inference, like other consistency models~\cite{luo2023latent,wang2023videolcm}, the classifier-free guidance~\cite{ho2022classifierfreeguidance} is not required, further reducing resource requirements.

\section{Experiments}

In this section, we will detail our experimental setup and present quantitative and qualitative results that demonstrate the effectiveness of our method.
\textbf{For more experiments and speed comparisons, please refer to the appendix}.

\begin{table*}[t]
%
%
\small
\setlength\tabcolsep{5.5pt}
\centering
\begin{tabular}{l|c|c|c|ccccc}
\shline
\hspace{-1.7mm}Method       & Steps & Diffusion model & Consistency model  & L1 $\downarrow$ & PSNR $\uparrow$ & SSIM $\uparrow$ & LPIPS $\downarrow$  & FVD $\downarrow$ \\ \shline
\hspace{-1.7mm}DreamPose~\cite{karras2023dreampose} 
 $_{\color{gray}{\text{(ICCV23)}}}$   & 100 & \Checkmark &
& 6.88E-04            & 12.82       & 0.511           & 0.442                            & 551.02           \\
\hspace{-1.7mm}DisCo~\cite{disco} 
 $_{\color{gray}{\text{(CVPR24)}}}$  & 50 & \Checkmark  &   
& 3.78E-04               & 16.55       & 0.668             & 0.292                                      & 292.80              \\
\hspace{-1.7mm}MagicAnimate~\cite{magicanimate} 
 $_{\color{gray}{\text{(CVPR24)}}}$ & 25 & \Checkmark &
& 3.13E-04     & -         & 0.714           & 0.239                                 & 179.07           \\
\hspace{-1.7mm}Animate Anyone~\cite{Animateanyone} 
 $_{\color{gray}{\text{(CVPR24)}}}$  & 20 & \Checkmark &
& -                & -       & 0.718             & 0.285                                            & 171.90              \\
\hspace{-1.7mm}Champ~\cite{champ} 
 $_{\color{gray}{\text{(ECCV24)}}}$ & 20 & \Checkmark &
& {2.94E-04}             & -     & {0.802}            & {0.234}                                        & {160.82}         \\
\hspace{-1.7mm}{UniAnimate}~\cite{wang2024unianimate} $_{\color{gray}{\text{(ArXiv24)}}}$ & 50  & \Checkmark & & \textbf{{2.66E-04}}      & \textbf{{20.58}}             & \textbf{{0.811}}            & \textbf{{0.231}}                                          & \textbf{{148.06}}         \\
\shline
\hspace{-1.7mm}{VideoLCM}~\cite{wang2023videolcm} $_{\color{gray}{\text{(ArXiv23)}}}$ & 4  &  & \Checkmark  &  2.85E-04      &         18.98      &       0.789       &       0.241                                     &196.21  \\
\hspace{-1.7mm}{PeRFlow}~\cite{yan2024perflow}
 $_{\color{gray}{\text{(NeurIPS24)}}}$ & 4 & &   & 2.59E-04      & 19.76             & 0.802          & 0.236                                          & 188.43  \\
\hspace{-1.7mm}{MultiStep CM}~\cite{heek2024multistep}
 $_{\color{gray}{\text{(ArXiv24)}}}$ & 4 & & \Checkmark  &   2.70E-04      &     19.39          &          0.802    &     0.231                                       &  187.59 \\
\rowcolor{Gray}
\hspace{-1.7mm}\textbf{\method} & 4 &  & \Checkmark &    \textbf{2.52E-04}    &        \textbf{20.12}       &        \textbf{0.812}        &                                     \textbf{0.225}     &     \textbf{156.27}     \\

\shline
\end{tabular} 
\vspace{-3mm}
\caption{
Quantitative comparisons with existing methods on TikTok.
The modified PSNR~\cite{wang2024unianimate,disco} is leveraged to avoid numerical overflow.
The accelerated baselines~\cite{wang2023videolcm,yan2024perflow,heek2024multistep} utilize the same teacher model and video data as our \method for distillation training.
}
\label{tab:quantitative_TikTok}
\vspace{1mm}
\end{table*}

\begin{table*}[t]
\small
\setlength\tabcolsep{6.5pt}
\centering
\begin{tabular}{l|c|c|c|cccc}
\shline
\hspace{-1.7mm}Method    & Steps      & Diffusion model & Consistency model & PSNR $\uparrow$ & SSIM $\uparrow$ & LPIPS $\downarrow$  & FVD $\downarrow$ \\ \shline

\hspace{-1.7mm}DreamPose~\cite{karras2023dreampose} $_{\color{gray}{\text{(ICCV23)}}}$ & 100  & \Checkmark &    &   - & 0.885 & 0.068  & 238.7        \\
\hspace{-1.7mm}DreamPose \emph{w/o Finetune}~\cite{karras2023dreampose} $_{\color{gray}{\text{(ICCV23)}}}$ & 100  & \Checkmark &   &   -  & 0.879 & 0.111 &  279.6        \\
\hspace{-1.7mm}Animate Anyone~\cite{Animateanyone} $_{\color{gray}{\text{(CVPR24)}}}$ & 20    & \Checkmark &  &   - & {0.931} & {0.044} & {81.6}             \\
\hspace{-1.7mm}{UniAnimate}~\cite{wang2024unianimate}
$_{\color{gray}{\text{(ArXiv24)}}}$  & 50  & \Checkmark &           &   \textbf{{27.56} }       & \textbf{{0.940} }           & \textbf{{0.031} }                                           & \textbf{{68.1}}         \\
\shline

\hspace{-1.7mm}{VideoLCM}~\cite{wang2023videolcm} $_{\color{gray}{\text{(ArXiv23)}}}$  & 4 &  &     \Checkmark      &   24.65        &    0.914         &                                      0.058       &     115.1     \\
\hspace{-1.7mm}{PeRFlow}~\cite{yan2024perflow}
 $_{\color{gray}{\text{(NeurIPS24)}}}$ & 4 &  &          &  26.28       &      0.922       &      0.044                                       &   89.7       \\
\hspace{-1.7mm}{MultiStep CM}~\cite{heek2024multistep}
 $_{\color{gray}{\text{(ArXiv24)}}}$ & 4 &  &     \Checkmark      &  25.81        &    0.929         &                                      0.047       &     88.0      \\
 \rowcolor{Gray}
\hspace{-1.7mm}{\textbf{\method}} & 4 &  &     \Checkmark      &   \textbf{27.54}       &      \textbf{0.938 }      &             \textbf{0.031 }                               &    \textbf{ 67.2}     \\

\shline
\end{tabular} 
\vspace{-3mm}
\caption{
Quantitative comparisons with existing methods on the UBC Fashion dataset.
The modified PSNR~\cite{wang2024unianimate,disco} metric is applied.
}
\label{tab:quantitative_fashion}
\vspace{-3mm}
\end{table*}

\begin{figure*}[t]
    \centering
    \includegraphics[width=1.0\linewidth]{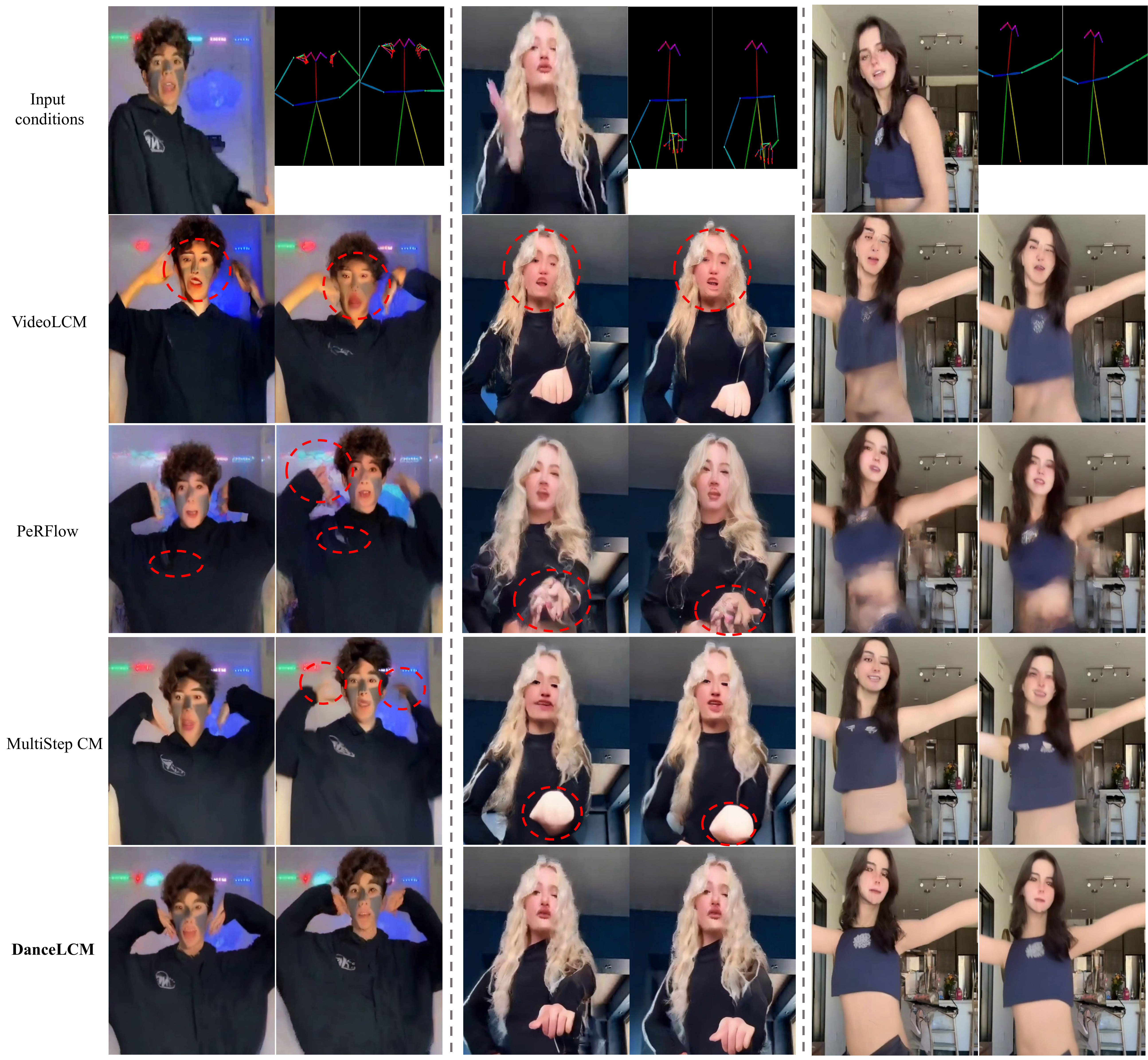}
    \vspace{-7mm}
    \caption{Qualitative evaluation on the TikTok dataset. Compared with the existing methods, the proposed \method achieves better results in terms of visual fidelity, temporal smoothness and face realism.
    }
    \label{fig:comparison}
    \vspace{-2mm}
\end{figure*}

\subsection{Experimental setup}

\noindent \textbf{Datasets and metrics.}
Following previous methods~\cite{Animateanyone,wang2024unianimate,magicanimate,champ,tan2024animatex}, we collected a training dataset with approximately 10K videos comprising various human dancing scenes.
To evaluate the performance, 
we employ the image metrics traditionally used in human image animation, including L1, Peak Signal-to-Noise Ratio (PSNR), Structural Similarity Index (SSIM), Learned Perceptual Image Patch Similarity (LPIPS), to quantitatively assess the visual fidelity of the generated videos. 
The video metric FVD~\cite{unterthiner2018towards} is also adopted.
The quantitative experiments are conducted on the test sets of TikTok~\cite{tiktokdata} and UBC Fashion~\cite{UBCfashion}.
%

\vspace{1mm}
\noindent \textbf{Implementation details.}
Following~\cite{Animateanyone,wang2024unianimate}, DWPose~\cite{DWpose} is leveraged to extract pose sequences of characters.
For the face condition, we use the MTCNN~\cite{MTCNN} algorithm to crop the face area and resize the image to 256$\times$256 spatial resolution.
We choose the state-of-the-art UniAnimate~\cite{wang2024unianimate} model as the teacher diffusion model to distillate the powerful knowledge.
For consistency distillation, we utilize the Adam optimizer with a learning rate of 2e-5 to update the student consistency model.
We uniformly sample $16/32$ frames at 768$\times$512 spatial resolution as input video to train the model. 
All experiments are conducted on 4 NVIDIA A100 GPUs.
If not specifically stated, \method generates video results with 4 inference denoising steps.


\subsection{Quantitative evaluation}
We present a comprehensive quantitative evaluation of \method against existing state-of-the-art methods on the TikTok and UBC Fashion datasets.
The results, as depicted in~\cref{tab:quantitative_TikTok} and~\cref{tab:quantitative_fashion}, showcase \method's superior performance over other competitive acceleration algorithms~\cite{wang2023videolcm,yan2024perflow,heek2024multistep} across all metrics under fair experimental settings.
%
\method achieves the lowest L1 error, indicating the closest match to the ground-truth, and the highest PSNR and SSIM values, signifying excellent visual quality and structural similarity. The lowest LPIPS and FVD scores further substantiate \method's ability to generate videos with high perceptual similarity and video quality.
In addition, we notice that our 4-step method is numerically comparable to the state-of-the-art UniAnimate~\cite{wang2024unianimate} with 50 DDIM steps, indicating that \method can accelerate inference while maintaining high performance.

\subsection{Qualitative evaluation}

To further evaluate the video quality, we conduct a qualitative comparison in~\cref{fig:comparison}.
The qualitative results showcase the visual quality and motion coherence of the videos generated by \method. 
The videos generated by \method exhibit sharper details, more natural motion transitions, and better preservation of facial features, particularly in dynamic regions.
This is evident when compared to the baseline VideoLCM, where our method significantly reduces motion blur and artifacts, offering a more lifelike animation quality.
Comparisons with other state-of-the-art methods highlight the effectiveness of our approach in preserving facial details and reducing motion artifacts.

\subsection{Ablation study}

\begin{figure*}[t]
    \centering
    \includegraphics[width=0.99\linewidth]{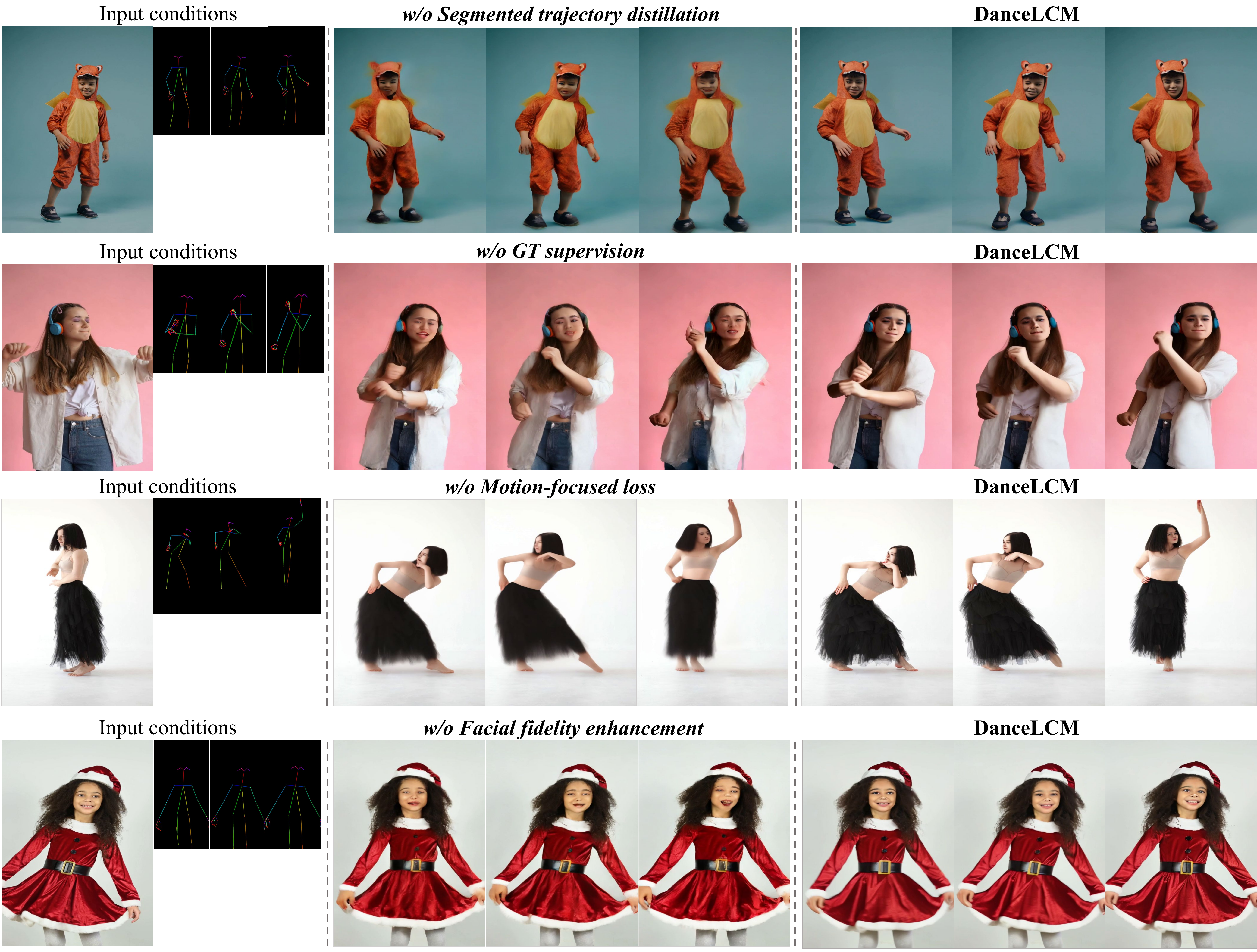}
    \vspace{-4mm}
    \caption{Qualitative ablation evaluation. Removing each component, the quality of the generated video decreases to some extent.}
    \label{fig:ablation}
    \vspace{-1mm}
\end{figure*}

\noindent \textbf{The efficacy of each module.
}
To dissect the contribution of each component within \method, we conduct a series of ablation studies, as detailed in~\cref{tab:ablation_study}. The removal of segmented distillation, GT supervision, motion-focused loss, and facial fidelity enhancement each result in a performance decrement, underscoring the integral role of these components in enhancing the overall quantitative performance of \method. 
Additionally, we conduct a qualitative evaluation in~\cref{fig:ablation} and further verify the effectiveness of each module.
Specifically, the segmented consistency distillation allows for more effective learning of clean spatial details.
GT supervision provides a reliable signal to guide the distillation process and helps to enhance sharp details.
The motion-focused loss ensures that the model pays adequate attention to dynamic regions of human movement, reducing motion blur. 
Lastly, the enhancement of facial fidelity ensures that the generated faces are realistic and consistent with the input conditions.

\begin{table}[t]
\small
\setlength\tabcolsep{2.0pt}
\centering
\begin{tabular}{l|cccc}
\shline
\hspace{-0.6mm}Method           & PSNR$\uparrow$ & SSIM$\uparrow$ & LPIPS$\downarrow$  & FVD$\downarrow$ \\ \shline

\hspace{-0.6mm}{{\method}}  &     \textbf{20.12}       &     \textbf{0.812}      &            \textbf{0.225}                               &     \textbf{156.27}     \\
\hspace{-0.6mm}{\emph{w/o} Segmented distillation}  &     19.48       &    0.803         &                                 0.228           &   169.23       \\
\hspace{-0.6mm}{\emph{w/o} GT supervision}  &     19.46       &      0.806       &            0.229                               &     161.63     \\
\hspace{-0.6mm}{\emph{w/o} Motion-focused loss}  &     19.56       &      0.809      &            0.227                               &    170.68     \\
\hspace{-0.6mm}{\emph{w/o} Facial fidelity enhancement}  &     19.77       &      0.811       & 0.231                                           &   166.92       \\
\shline
\end{tabular} 
\vspace{-3mm}
\caption{
Quantitative ablation evaluation on the TikTok dataset.
}
\label{tab:ablation_study}
\vspace{-3mm}
\end{table}

\begin{table}[t]
\small
\setlength\tabcolsep{3.8pt}
\centering
\begin{tabular}{l|c|cccc}
\shline
\hspace{-1.2mm}Method    & Steps       & PSNR $\uparrow$ & SSIM $\uparrow$ & LPIPS $\downarrow$  & FVD $\downarrow$ \\ \shline

\hspace{-1.2mm}{VideoLCM}~\cite{wang2023videolcm} & 8 &     19.03       &   0.793    &    0.244                                     &    179.33    \\

\hspace{-1.2mm}{{\method}} & 8 &  \textbf{19.73}   &    \textbf{0.814}    &     \textbf{0.221}                                    &   \textbf{152.64}     \\
\hline
\hspace{-1.2mm}{VideoLCM}~\cite{wang2023videolcm} & 2 &      18.62      &    0.789    &     0.253                                    &     207.72   \\

\hspace{-1.2mm}{{\method}} & 2 &  \textbf{19.36}   &    \textbf{0.807}    &     \textbf{0.228}                                    &   \textbf{188.73}     \\
\hline

\hspace{-1.2mm}{VideoLCM}~\cite{wang2023videolcm} & 1 &         18.30   &     0.778   &                                0.257         &    224.78     \\

\hspace{-1.2mm}{{\method}} & 1 &   \textbf{19.17}         &     \textbf{0.798}   &                                \textbf{0.239}         &    \textbf{190.22}    \\

\shline
\end{tabular} 
\vspace{-3mm}
\caption{
Ablation study on the number of inference steps.
}
\label{tab:ablation_study_steps}
\end{table}

\begin{figure}
    \centering
    \includegraphics[width=1.0\linewidth]{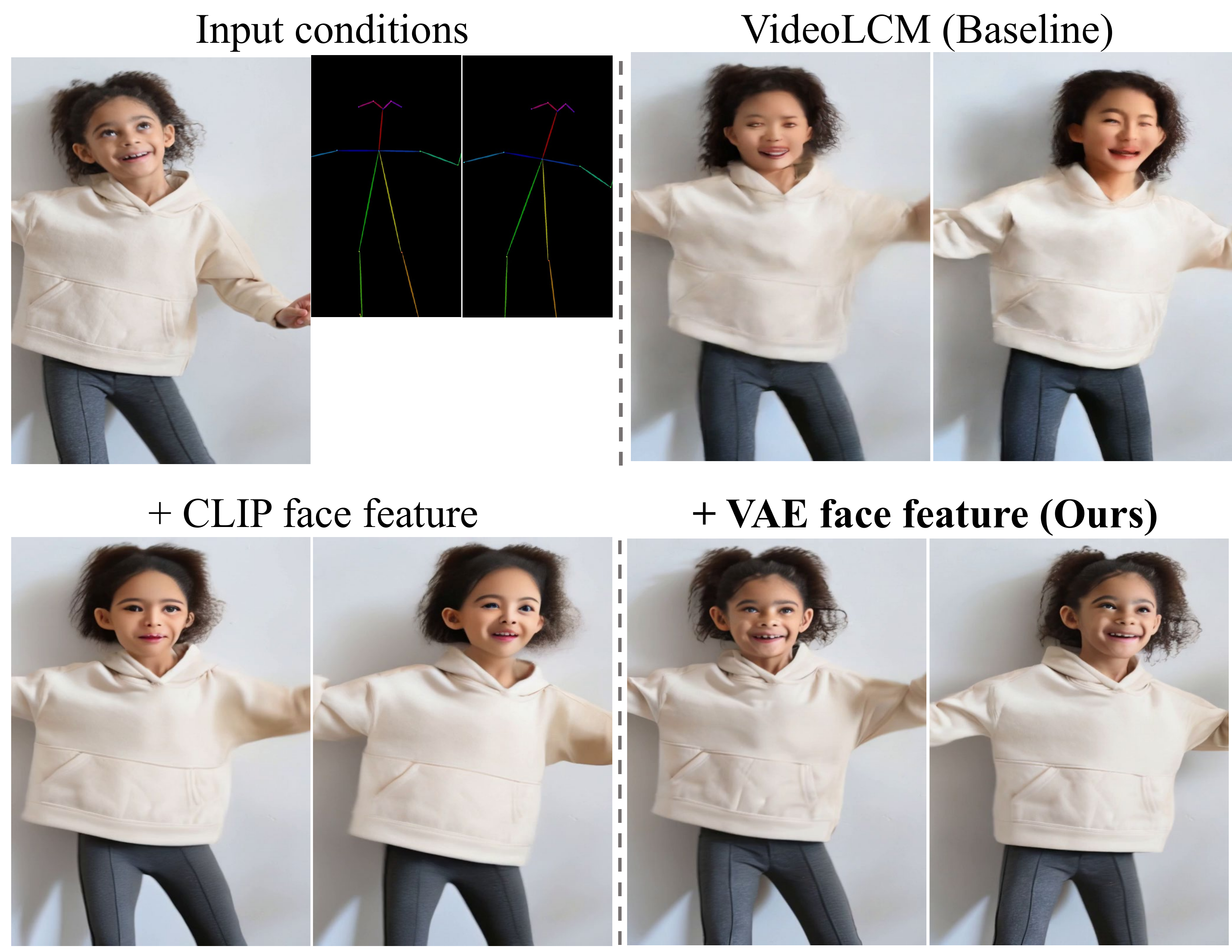}
    \vspace{-7mm}
    \caption{Ablation study on effect of facial fidelity enhancement.}
    \label{fig:face_ablation}
    \vspace{-4mm}
\end{figure}

\begin{table}[t]
\small
\setlength\tabcolsep{4.5pt}
\centering
\begin{tabular}{l|cccc}
\shline
\hspace{-1.5mm}Number of segments           & PSNR $\uparrow$ & SSIM $\uparrow$ & LPIPS $\downarrow$  & FVD $\downarrow$ \\ \shline

\hspace{-1.5mm}{$K$ = 1}  &    19.20       &      0.801       &             0.232                               &     186.30     \\
\hspace{-1.5mm}{$K$ = 2}  &   \textbf{20.12}       &      0.812       &            \textbf{0.225}                               &     \textbf{156.27}     \\   
\hspace{-1.5mm}{$K$ = 4 }  &     20.23       &      \textbf{0.813}       &             0.229                               &    166.92    \\
\shline
\end{tabular} 
\vspace{-3mm}
\caption{
Ablation study on the number of segments in the segmented consistency distillation with 4 inference steps.
}
\label{tab:ablation_study_segments}
\vspace{-5mm}
\end{table}

\begin{figure*}[t]
    \centering
    \includegraphics[width=1.0\linewidth]{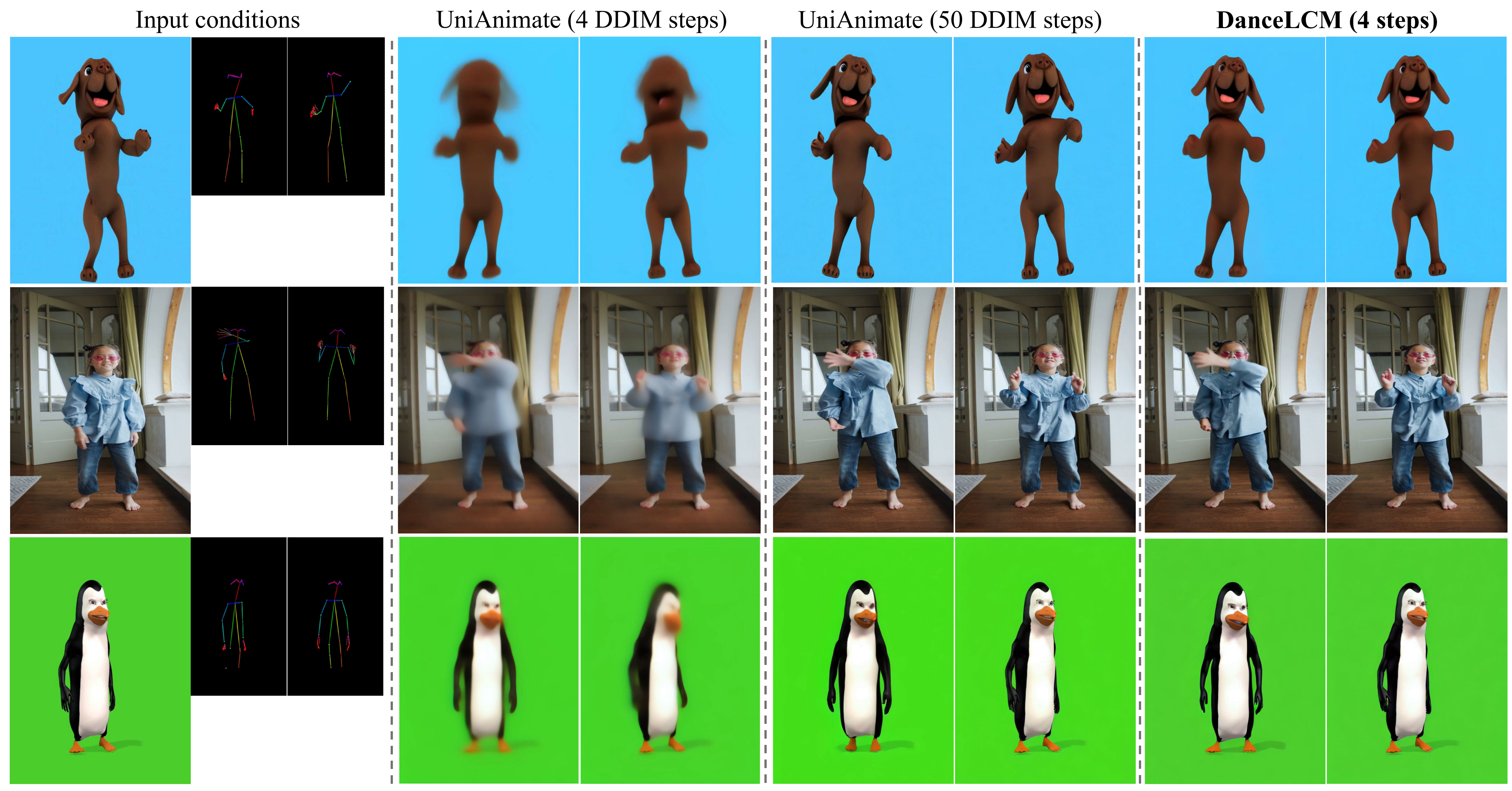}
    \vspace{-8mm}
    \caption{Qualitative comparison with teacher models. \method with 4 inference steps can achieve a significant lead over the 4-step UniAnimate results in terms of visual quality and is comparable to the 50-step UniAnimate results.}
    \label{fig:comparison_with_teacher}
\end{figure*}

\vspace{1mm}
\noindent \textbf{The impact of VAE face feature.
}
To enhance the facial fidelity, we incorporate the VAE face feature in our method. 
%
%
The VAE is utilized to extract rich and detailed facial features from the reference face image, ensuring that the generated faces maintain realistic attributes.
To investigate the effectiveness of incorporating the VAE face feature, we conduct a qualitative comparison with the CLIP face feature.
As displayed in~\cref{fig:face_ablation}, 
the CLIP face feature tends to capture global semantic information but may overlook finer facial details. In contrast, the integration of VAE facial feature can significantly improve the realism of facial animations in the generated videos.
The above observations verify the rationality of our design.

\vspace{1mm}
\noindent \textbf{Varying the number of inference steps.
}
We also explore the impact of varying the number of inference steps on the performance of \method. 
As shown in~\cref{tab:ablation_study_steps}, we notice that the performance will improve with the increase of the number of steps, and \method achieves consistently better quantitative performance than the VideoLCM baseline.



\vspace{1mm}
\noindent \textbf{Varying the number of segments.
}
In our \method, segmented consistency distillation is performed.
In~\cref{tab:ablation_study_segments}, we further analyze the effect of different number of segments in segmented consistency distillation.
The number of inference steps is kept at 4. 
We can notice that $K=2$ achieves a good balance.
We attribute this phenomenon to the fact that $K=1$ leads to high learning difficulty, as the model requires to map any points on the entire PF-ODE trajectory to the start point. 
When $K=4$,
there may be a certain amount of error within each segment, and too many segments will lead to error accumulation.

\subsection{Qualitative comparison with teacher models}

In \cref{fig:comparison_with_teacher}, we present a qualitative comparison between \method and the teacher diffusion model  at different inference steps. The results indicate that the proposed \method can achieve significant improvements over the baseline teacher model with 4 steps, and is even comparable to the results of the teacher model with a much higher number of steps in terms of visual quality and temporal coherence. This comparison underscores the effectiveness of \method in accelerating the  animation process without compromising video quality of the generated content.

\section{Conclusion and limitations}

In this paper, we introduced the \method framework, accelerating existing diffusion models in generating high-quality, temporally coherent videos for human image animation task. 
Specifically,
we propose a tailored consistency distillation methodology that employs segmented trajectory distillation with ground-truth supervision, enabling more effective learning of spatial details while minimizing error accumulation. 
Additionally, we enhance visual fidelity and motion continuity by emphasizing key motion areas and integrating facial fidelity features, leading to superior high-quality outcomes.
Both qualitative and quantitative evaluations verify the effectiveness of the proposed \method.
%

\vspace{0.2mm}
\noindent \textbf{Limitations.}
While 
\method demonstrates significant advancements in accelerated human image animation, several limitations remain. First, the fidelity of intricate details, such as fingers, is still suboptimal. 
Second, the generation results of our approach are reliant on the performance of the teacher model.
Future work will focus on enhancing the representation of delicate features, such as fingers, and improving the performance of the teacher model to ensure more accurate and high-fidelity animations.
%


\section*{Acknowledgment}
This work is supported by the National Natural Science Foundation
of China under grants U22B2053 and 623B2039, and Alibaba Group through Alibaba Research Intern Program.

{
\small
\bibliographystyle{ieeenat_fullname}
\bibliography{ref.bib}
}


\end{document}